# In the Service of Online Order: Tackling Cyber-Bullying with Machine Learning and Affect Analysis

Michal Ptaszynski [1]  Pawel Dybala [2]  Tatsuaki Matsuba [3]  Fumito Masui [4]  Rafal Rzepka [2]

Kenji Araki [2]  Yoshio Momouchi [5]

**Abstract.** One of the burning problems lately in Japan has been cyber-bullying, or slandering and bullying people online. The problem has been especially noticed on unofficial Web sites of Japanese schools. Volunteers consisting of school personnel and PTA (Parent-Teacher Association) members have started Online Patrol to spot malicious contents within Web forums and blogs. In practise, Online Patrol assumes reading through the whole Web contents, which is a task difficult to perform manually. With this paper we introduce a research intended to help PTA members perform Online Patrol more efficiently. We aim to develop a set of tools that can automatically detect malicious entries and report them to PTA members. First, we collected cyber-bullying data from unofficial school Web sites. Then we performed analysis of this data in two ways. Firstly, we analysed the entries with a multifaceted affect analysis system in order to find distinctive features for cyber-bullying and apply them to a machine learning classifier. Secondly, we applied a SVM based machine learning method to train a classifier for detection of cyber-bullying. The system was able to classify cyber-bullying entries with 88.2% of balanced F-score.

## 1 INTRODUCTION

Online security has been an urgent problem ever since the creation of the Internet. Online security is a sub-field of information security applied to Internet and networks. It deals with all sorts of undesirable entities or events appearing on the Internet. Some of the most well known issues of online security include hacking, cracking, data theft and online espionage. However, for several years, a problem that has become much more visible and therefore influential and socially harmful, is the problem of exploitation of online open communication means, such as BBS forum boards, or social networks to convey harmful and disturbing information. In USA, a great focus on this issue began in 2001 after the 9.11 terrorist attack. However, similar cases have been noticed before in other countries on a smaller scale. In Japan, on which this research is focused, a great social disturbance was caused by cases of sending alarming messages by criminals or hijackers on the Internet just before committing a crime. One famous case of this kind happened in May 2000, a year before the 9.11 in USA, when a frustrated young man sent a message on a popular Japanese BBS forum *2channel*, informing readers he was going to hijack a bus, just before he proceeded with his plan. Growing number of similar cases around the world opened a public debate on whether such suspicious messages could not be spotted early enough to prevent the crimes from happening [1] and on the freedom of speech on the Internet in general [2]. Some of the famous research in this matter was done by the team of Hsinchun Chen, who started the project aiming to analyze Dark Web Forums in search of alarming entries about planned terrorist attacks [3, 4]. Another research of this kind was performed by Gerstenfeld [5], who focused on extremist groups.

However, there have been little research performed on a problem less lethal, although equally serious, namely online slandering and bullying of private persons, known generally as 'cyber-bullying'. In Japan the problem has become serious enough to be noticed by the Ministry of Education, Culture, Sports, Science and Technology (later: MEXT) [6]. At present, school personnel and members of Parent-Teacher Association (PTA) have started Online Patrol to spot Web sites and blogs containing such inappropriate contents. However, countless number of such data makes the job an uphill task. Moreover, the Online Patrol is performed manually and as a volunteer work. Therefore we started this research to help the Online Patrol members. The final goal of this research is to create a machine Online Patrol crawler that can automatically spot the cyber-bullying cases on the Web and report them to the Police. In this paper we present some of the first results of this research. We first focused on developing a systematic approach to spotting online cyber-bullying entries automatically to ease the burden of the Online Patrol volunteers. In our approach we perform affect analysis of these contents to find distinctive features for cyber-bullying entries. The feature specified as the most characteristic for cyber-bullying is used in training of a machine learning algorithm for spotting the malicious contents.

The paper outline is as follows. In Section 2 we describe the problem of cyber-bullying in more details. In Section 3 we present the detailed description of affect analysis systems used in this research. Section 4 contains results of affect analysis of cyber-bullying data and propose a feature to apply in training of machine learning algorithm. In Section 5 we present the prototype SVM-based method for detecting the cyber-bullying entries and in the Section 6 evaluate it. Finally, in Section 7 we conclude the paper and provide some hints on further work in this area.

[1] JSPS Research Fellow / High-Tech Research Center, Hokkai-Gakuen University, ptaszynski@hgu.jp
[2] Graduate School of Information Science and Technology, Hokkaido University, {paweldybala,kabura,araki}@media.eng.hokudai.ac.jp
[3] Graduate School of Engineering, Mie University, matsuba@ai.info.mie-u.ac.jp
[4] Department of Computer Science, Kitami Institute of Technology, f-masui@mail.kitami-it.ac.jp
[5] Department of Electronics and Information Engineering, Faculty of Engineering, Hokkai-Gakuen University, momouchi@eli.hokkai-s-u.ac.jp

## 2 WHAT IS CYBER-BULLYING?

Although the problem of sending harmful messages on the Internet has existed for several years, it has been officially defined only recently and named as cyber-bullying[6]. The National Crime Prevention Council in USA state that cyber-bullying happens "when the Internet, cell phones or other devices are used to send or post text or images intended to hurt or embarrass another person."[7]. Other definitions, such as the one by Bill Belsey, a teacher and an anti-cyber-bullying activist, say that cyber-bullying "involves the use of information and communication technologies to support deliberate, repeated, and hostile behaviour by an individual or group, that is intended to harm others." [7].

Some of the first robust research on cyber-bullying was done by Hinduja and Patchin, who performed numerous surveys about the subject in the USA [8, 9]. They found out that the harmful information may include threats, sexual remarks, pejorative labels, false statements aimed at humiliation. When posted on a BBS forum or a social network, such as Facebook, it may disclose personal data of the victim. The data which contains humiliating information about the victim defames or ridicules the victim personally.

### 2.1 Cyber-bullying and Online Patrol in Japan

In Japan, after a several cases of suicides of cyber-bullying victims who could not bare the humiliation, MEXT has considered the problem serious enough to start a movement against the problem. In a manual for spotting and handling the cases of cyber-bullying [6], the Ministry puts a great importance on early spotting of the suspicious entries and messages, and distinguishes several types of cyber-bullying noticed in Japan. These are:

1. Cyber-bullying appearing on BBS forums, blogs and on private profile web-sites;
   (a) Entries containing libelous, slanderous or abusive contents;
   (b) Disclosing personal data of natural persons without their authorization;
   (c) Entries and humiliating online activities performed in the name of another person;
2. Cyber-bullying appearing in electronic mail;
   (a) E-mails directed to a certain person/child, containing libelous, slanderous or abusive contents;
   (b) E-mails in the form of chain letters containing libelous, slanderous or abusive contents;
   (c) E-mails send in the name of another person, containing humiliating contents;

In this research we focused mostly on the cases of cyber-bullying that appear on informal web sites of Japanese secondary schools. Informal school web sites are web sites where school pupils gather to exchange information about school subjects or contents of tests, etc. However, as was noticed by Watanabe and Sunayama [10], on such pages there have been a rapid increase of entries containing insulting or slandering information about other pupils or even teachers. Cases like that make other users uncomfortable using the Web sites and cause undesirable misunderstandings.

---
[6] other terms used are cyber-harassment, and cyber-stalking
[7] http://www.ncpc.org/cyberbullying

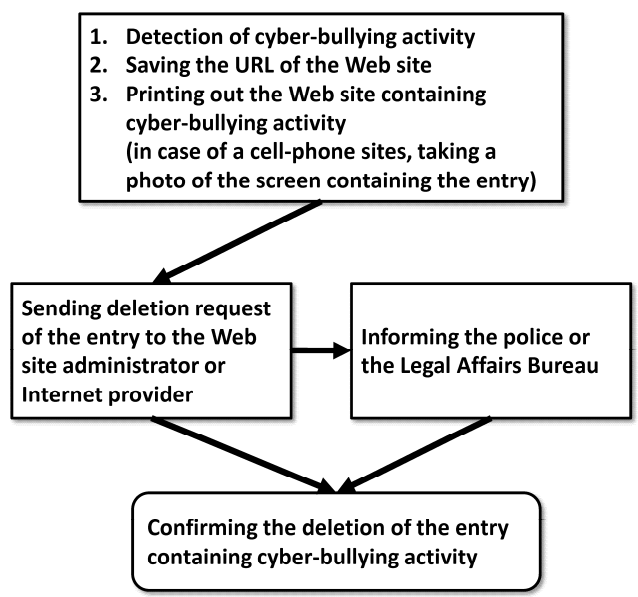

**Figure 1.** Route of Online Patrol.

To deal with such malicious entries, a movement of Online Patrol (OP) was founded. Participants of this movement are usually teachers and PTA members. Based on the MEXT definition of cyber-bullying, they read through all available entries, decide whether an entry is dangerous or not and, if necessary, send a deletion request to the web page administrator. Finally, they send a report about the event to the police. The typical Online Patrol route is presented on Figure 1.

Unfortunately, at present state of affairs, the school personnel and PTA members taking part in Online Patrol perform all tasks manually as voluntary work, beginning from reading the countless numbers of entries and deciding about the appropriateness of their contents, through printing out or taking photos of the pages containing cyber-bullying entries, and finally sending the reports and deletion requests to the appropriate organs. Moreover, since the number of entries rises by the day, surveillance of the whole Web becomes an uphill task for the small number of patrol members.

With this research we aim to create a Web crawler capable to perform this difficult task instead of humans, or at least to ease the burden of the Online Patrol volunteers.

## 3 AFFECT ANALYSIS OF CYBER-BULLYING

As was shown by Chen and colleagues [3, 4], analysis of affect intensity of Dark Web Forums often helps specifying the character of the forum. Aiming to find any dependencies between expressing emotions and cyber-bullying activities we performed a study and analysed affective level of the cyber-bullying data.

For contrastive affect analysis we obtained Online Patrol data containing both cyber-bullying activities and normal entries. The data used in affect analysis contained 1,495 harmful and 1,504 non-harmful entries. The affect analysis was performed with ML-Ask system for affect analysis of textual input in Japanese and CAO system for analysis of emoticons, both developed by Ptaszynski and colleagues [16, 18].

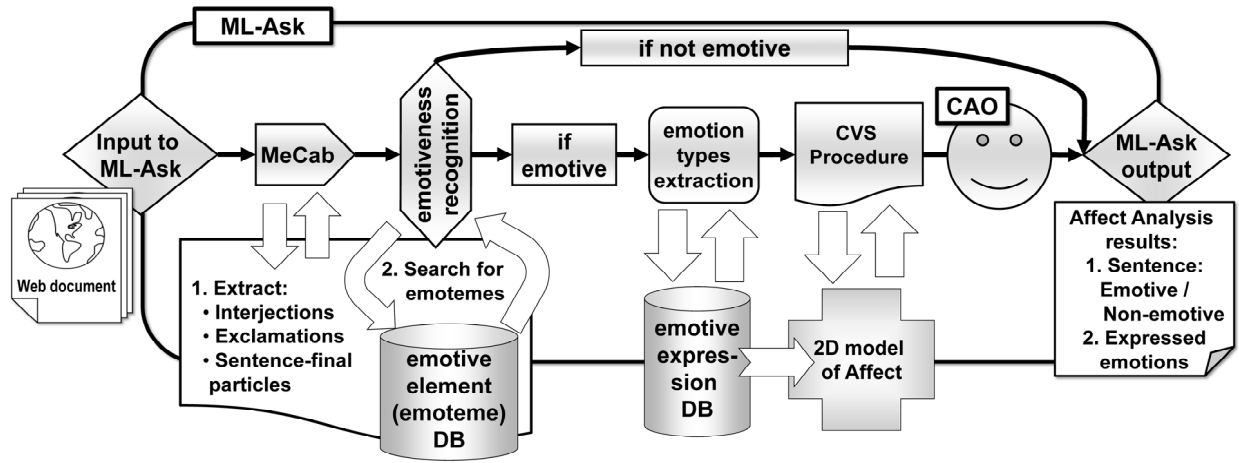

**Figure 2.** Flow chart of ML-Ask system.

## 3.1 ML-Ask - Affect Analysis System

ML-Ask (eMotive eLements / Emotive Expressions Analysis System) is a system developed for analyzing the emotive contents of utterances. It provides not only information on emotions expressed in input, but also linguistic information on what sentence elements represent which emotive features and what is their grammatical classification. The system uses a two-step procedure: 1) Analyzing the general emotiveness of an utterance by detecting emotive elements, or emotemes, expressed by the speaker and classifying the utterance as emotive or non-emotive; 2) Recognizing the particular emotion types by extracting expressions of particular emotions from the utterance. This analysis is based on Ptaszynski's [21] idea of two-part classification of realizations of emotions in language into:

**1) Emotive elements or Emotemes.** Elements conveyed in an utterance indicating that the speaker was emotionally involved in the utterance, but not detailing the specific emotions. The same emotive element can express different emotions depending on context. This group is linguistically realized by subgroups such as interjections, exclamations, mimetic expressions, or vulgar language. Examples are: *sugee* (great!), *wakuwaku* (heart pounding), *-yagaru* (a vulgarization of a verb);

**2) Emotive expressions.** Words used to describe emotional states. However, they function as expressions of the speaker's emotions only in utterances where the speaker is emotionally involved. In non-emotive sentences they fulfill the function of simple descriptive expressions. The group is realized by various parts of speech, like nouns, verbs, adjectives, etc. Examples are: *aijou* (love), *kanashimu* (feel sad), *ureshii* (happy), respectively.

The manually-selected emotive element database consists of interjections, mimetic expressions, endearments, vulgarities, and representations of non-verbal emotive elements, such as exclamation marks or ellipses. The emoteme database collected and divided in this way contains 907 elements in total.

A system for analysis of emoticons was also added, as emoticons are symbols commonly used in everyday text-based communication to convey emotions. The description of the system for emoticon analysis is presented in section 3.2.

The database of emotive expressions is based on Nakamura's collection [22] and contains 2100 emotive expressions, each classified into the emotion type they express.

### 3.1.1 Affect Analysis Procedure

On textual input provided by the user, two features are computed in order: the emotiveness of an utterance and the specific type of emotion. To determine the first feature, the system searches for emotive elements in the input to determine whether the input is emotive or non-emotive. In order to do this, the system uses MeCab [17] for morphological analysis and separates every part of speech. MeCab recognizes some parts of speech which belong to the group of emotemes, such as interjections, exclamations or emphatic sentence-final particles, like *-zo*, *-yo*, or *-ne*. If these appear, they are extracted from the utterance as emotemes. Next, the system searches and extracts every emoteme based on the system's emoteme databases (907 items). The input is then processed by CAO [18], a system for analysis of emoticons (described in detail in section 3.2). Emoticons belong to both emotemes, and emotive expressions, as their appearance in an input always indicates emotional attitude of the user and for most emoticons it is possible to specify the emotion type they convey. All of the extracted elements mentioned above (exclamations from MeCab, emotemes and emoticons) indicate the emotional level of the utterance. The emotional value of an input is calculated quantitatively; the more emotemes, the higher is the emotional value. The upper limit is set as 5.

Secondly, in utterances classified as emotive (presence of at least one emoteme), the system uses a database of emotive expressions to search for all expressions describing emotional states. Here, emotions specified by CAO are added to the result of ML-Ask basic procedure. This determines the specific emotion type (or types) conveyed in the utterance. Examples of analysis performed by ML-Ask (system output) are represented below after Section 3.2. In these examples, from top line there are: an example in Japanese, emotive information annotation (emotemes-underlined, emotive expressions-bold type font) and English translation.

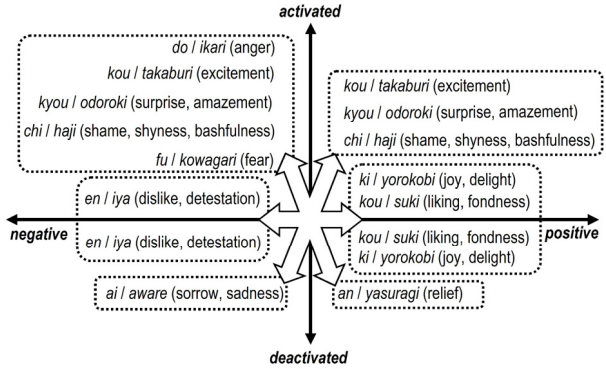

**Figure 3.** Grouping Nakamura's classification of emotions on Russell's space.

*3.1.2 Contextual Valence Shifters in ML-Ask*

One of the problems in the procedure described above was confusing in some cases the valence polarity of emotive expressions. The cause of this problem was extracting from the utterance only the emotive expression keywords without the grammatical context. One case of such an input is presented below in example (3). In this sentence the emotive expression is the verb *akirameru* (to give up [verb]) but the phrase *-cha ikenai* (Don't- [particle+verb]) suggest that the speaker is in fact negating and forbidding the emotion expressed literally. Such phrases are called Contextual Valence Shifters (CVS).

The idea of Contextual Valence Shifters (CVS) application in Sentiment Analysis was first proposed by Polanyi and Zaenen [23]. They distinguished two kinds of CVS: negations and intensifiers. The group of negations contains words and phrases like "not", "never", and "not quite", which change the valence polarity of the semantic orientation of an evaluative word they are attached to. The group of intensifiers contains words like "very", "very much", and "deeply", which intensify the semantic orientation of an evaluative word. Examples of CVS negations in Japanese are grammatical structures such as: *-nai* (not-), *amari -nai* (not quite-), *-to wa ienai* (cannot say it is-), or *-te wa ikenai* (cannot+[verb]-). Intensifiers are: *totemo-* (very-), *sugoku-* (-a lot), or *kiwamete-* (extremely). In this research we focused mostly on negations, since they have an immediate and significant influence on the meaning of emotive expressions. We applied Contextual Valence Shifters to change the valence polarity of emotive expressions in utterances containing CVS structures. Our manually-crafted database of CVS contains 71 negation structures.

However, using only the CVS analysis we would be able to find out the appropriate valence of emotions conveyed in the utterance, but we would not know the exact emotion type. To specify the emotion types after changing polarity with CVS, we applied the idea of the 2-dimensional model of affect [24] which assumes that all emotions can be described in 2-dimensions: the emotion's valence polarity (positive/negative) and activation (activated/deactivated). An example of positive-activated emotion could be "excitement"; a positive-deactivated emotion is, e.g., "relief" (see Figure 3).

Emotion types distinguished by Nakamura [22] were mapped on this model and their affiliation to one of the spaces determined. The emotion types with ambiguous affiliation were mapped on two possible fields. When a CVS structure is discovered, ML-Ask changes the valence polarity of the detected emotion. The appropriate emotion after valence changing is determined as the one with valence polarity and activation parameters different to the contrasted emotion (note arrows in Figure 3).

### 3.2 CAO - Emoticon Analysis System

Since one of the most popular strategies of expressing emotions in online communication is using emoticons, we supported ML-Ask with emoticon analysis system CAO. It is a system for estimation of emotions conveyed through emoticons reported by Ptaszynski and colleagues [18]. Emoticons are - sets of symbols widely used in text-based online communication to convey emotions. CAO, or *emotiCon Analysis and decOding of affective information*, extracts an emoticon form an input (a sentence) and determines specific emotion types expressed by it using a three-step procedure. Firstly, matching the input with a predetermined raw emoticon database containing over ten thousand emoticons. The emoticons, which could not be estimated with only the database are automatically divided into semantic areas, such as representations of "mouth" or "eyes", based on the idea of *kinemes*, or minimal meaningful body movements, applied from the theory of kinesics [19, 20]. The areas are automatically annotated according to their co-occurrence in database. The annotation is firstly based on eye-mouth-eye triplet. If no triplet was found, all semantic areas are estimated separately. This provides hints about potential groups of expressed emotions giving the system a coverage of over 3 million possibilities. CAO is used as a supporting procedure in ML-Ask to improve performance of the affect analysis system in utterances, which do not include emotive expressions, like in the example (2) below (emotemes - underlined; emotive expressions - bold type font).

(1) *Kyo wa   nante   kimochi ii   hi   nanda   !*
Today:TOP   ML:nante **MX:joy**   day:SUB   ML:nanda   ML:!
Translation: Today is such a nice day!

(2) *Iya~,   sore wa   sugoi   desu   ne–   !   ^o^*
ML:iya~  this  :TOP   ML:sugoi   COP   ML:ne–   ML:!   ML:^o^
Translation: Whoa, that's great! ^o^   **MX:joy**

(3) **Akirame**   *cha   ikenai*   *yo*   *!*
**MX:dislike**   ML:cha | CVS:cha-ikenai{→joy}   ML:yo   ML:!
Translation: Don't give up!

(4) **Hitoribocchi**   *nante*   *iya*   *da*   *~~*
**MX:sadness**   ML:nante–da **MX:dislike**   COP   ML:~~
Translation: Being alone sucks...

## 4 AFFECT ANALYSIS RESULTS

At first we calculated the number of all emotive entries among both sets of data. There was 956 emotive samples among 1,495 harmful (63.95%) and 1,029 among 1,504 non-harmful (68.42%) entries. The difference was not high and therefore the number of emotive entries cannot be considered as a highly distinctive feature. However, we made the first assumption, that harmful data are less emotively emphasized than non-harmful. This is a reasonable assumption, since cyber-bullying is often based on irony or sarcasm, which is not highly emotive, however deliberately uses some amount of emotive information to slander the object of sarcasm. To confirm this thesis we performed other comparisons.

We calculated emotive values of all emotive entries. Although the number of emotive entries and emotive value, both relate to the idea of emotiveness of a corpus in general, they are not directly related. One can easily imagine the difference between one corpus that consists of many slightly emotive entries (high number of emotive entries, but low average emotive value) and another corpus with a small number of highly emotive entries (low number of emotive entries, but high average emotive value). The approximated emotive value for harmful and non-harmful data was 1.47 and 1.5, respectively. Here also the difference is not high, although, since both values are not directly related, this can moderately support the thesis.

**Table 1.** Four examples of cyber-bullying entries gathered during Online Patrol from the unofficial school BBS Web site. The upper three represent expressing strong sarcasm despite of the use of positive expressions in the sentence. Emotemes - underlined; Emotive expressions - bolt type font.

| |
|---|
| *>>104 Senzuri koite shinu nante, sonna hageshii senzuri sugee naa. "Senzuri masutaa" toshite isshou **agamete** yaru yo.* <br> Analysis results: Emotemes: *nante, sugee, naa, -yo, senzuri*; Emotive expression: **agameru** (worship, respect) → **fondness** ; <br> Translation: >>104 Dying by 'flicking the bean'? Cannot imagine how one could 'flick the bean' so fiercely. I'll worship you forever, as a 'master-bator'. |
| *2-nen no tsutsuji no onna meccha busu **suki** na hito barashimashouka? 1-nen no ano ko desu yo ne? kimogatteru n de yamete agete kudasai* <br> Analysis results: Emotemes: *meccha, -yo, -ne, -nde, kimogaru*; Emotive expressions: **suki** (like) → **fondness** ; <br> Translation: Ya wanna know who likes 2nd-grade ugly azalea girls? Its that 1st-grader isn't it? He's looks disgusted, so leave him mercifully in peace. |
| *Aitsu wa busakute se ga takai dake no onna, busakute se takai dake ya no ni yatara otoko-**zuki** meccha tarashi de panko anna onna owatteru* <br> Analysis results: Emotemes: *yatara, meccha, busai, tarashi, panko*; Emotive expressions: **-zuki** (-lover; an amateur of-) → **fondness** ; <br> Translation: She's just tall and apart of that she's so freakin' ugly, and despite of that she's such a cock-loving slut, she's finished already. |
| *Shinde kureee, daibu **kiraware-mono** de yuumei, subete ga **itaitashii**... |* Analysis results: Emotemes: *–eee* (syllable prolongation), *...* (elipsis); <br> Emotive expressions: **kiraware-mono** (disliked by others) → **dislike**, ***itaitashii*** (pathetic, pitiful) → **gloom, sadness** ; <br> Translation: Please, dieee, you're so famous for being disliked by everyone, everything in you is so pathetic |

Next, we took a closer look on the extracted emotemes. There are four groups of emotemes distinguished in ML-Ask: i) interjections, ii) exclamations, iii) vulgarities and iv) mimetic expressions (*gitaigo* in Japanese) - arranged in order of their emotional weight. Distribution of the extracted emotemes within both entry sets is represented in Table 2. There were relatively more emotemes with high emotive weight in non-harmful data and more low weight emotemes in harmful data, which is another confirmation of the thesis presented at the beginning of this section.

The biggest difference between the number of extracted emotemes was found for vulgarities, which can be regarded as a unique and distinctive feature of cyber-bullying entries. This is one of the reasons we used it as the main feature in machine learning system, described later in section 5.

A similar difference, although lower, appeared also in mimetic expressions, which could be included in further study on enlarging lexicon for the machine learning system. As for interjections and exclamations, although they appeared in a large number in both datasets, more of them appeared in non-harmful entries. This could be caused by the fact that these two emoteme types are used to express emotive attitude in a straightforward way. Therefore, although there certainly are cyber-bullying cases where the victims are slandered straightforwardly, not emotional and cold sarcasm is also an often phenomenon.

**Table 2.** Distribution of the extracted emotemes within both entry sets.

| type of | type of data | |
|---|---|---|
| emoteme | non-harmful | harmful |
| interjections | 859 | 784 |
| exclamations | 284 | 174 |
| vulgarities | 8 | **149** |
| mimetic expressions | 7 | **23** |

Another comparison was made with the closer study of emotive utterances. ML-Ask is capable of detecting emotive utterances with a high reliability (*kappa* = 0.8), although it specifies particular emotion types with low Recall (although with high Precision) [16]. It is due to the use of a lexicon (Nakamura's dictionary, cf. [22]) that is out of date. Therefore there is a certain number of samples always described as emotive but with no specified emotion types. Such a phenomenon is however reasonable from the linguistic point of view, since there are many sentences that are emotive, although the emotion they convey depends on their context.

We first compared how many there were specified vs unspecified emotive utterances. The result was 13.18% vs 86.82% for cyber-bullying and 11.95% vs 88.05% for the normal entries. The higher ratio of specified emotion types in cyber-bullying data might suggest that people more often use traditional emotive expressions to slander people than they do to express emotions usually.

**Table 3.** The number of particular emotive expressions extracted by ML-Ask from both datasets.

| type of data | | | |
|---|---|---|---|
| non-harmful | | harmful | |
| emotion type | no. extr. | emotion type | no. extr. |
| dislike | 49 | dislike | 56 |
| joy | 32 | fondness | 49 |
| fondness | 21 | joy | 12 |
| relief | 18 | relief | 9 |
| fear | 11 | anger | 8 |
| gloom/sadness | 7 | gloom/sadness | 7 |
| surprize | 6 | excitement | 3 |
| excitement | 5 | fear | 3 |
| anger | 3 | shame | 3 |
| shame | 0 | surprize | 3 |

Next, we compared the number of particular emotive expressions extracted by ML-Ask. The results have been represented in Table 3. The analysis of the number of emotion types represented by the extracted emotive expressions revealed interesting tendencies, although here as well the strength of the proof is low. However, anger scored much higher in the cyber-bullying data than in the normal data, which is a reasonable result, since Web site entries meant to slander others are expected to express anger more often. Fear on the other hand scored very low in cyber-bullying data. This is also reasonable, since it is difficult to bully others by expressing one's fears. Dislike scored slightly higher in the harmful data. As for the positive emotion types, e.g., joy scored higher in normal dataset, which was expectable.

**Table 4.** Comparison of tendencies in annotation of emotion types with regard to the two-dimensional affect space.

| valence (polarity) of emotions | | | | | | | | |
|---|---|---|---|---|---|---|---|---|
| negative emotion | | | negative/positive (both possible) | | | positive emotion | | |
| emotion type | non-harmful | harmful | emotion type | non-harmful | harmful | emotion type | non-harmful | harmful |
| gloom/sadness | 7 | 7 | excitement | 5 | 3 | joy | **32** | 12 |
| fear | **11** | 3 | shame | 0 | **3** | fondness | 21 | **49** |
| anger | 3 | **8** | surprize | **6** | 3 | relief | **18** | 9 |
| dislike | 49 | **56** | | | | SUM | **71** | 70 |
| SUM | 70 | **74** | SUM | **11** | 9 | SUM (fondness excluded) | **50** | 21 |

| activation of emotions | | | | | | | | |
|---|---|---|---|---|---|---|---|---|
| deactivated emotion | | | moderately activated (deactivated/activated) | | | activated emotion | | |
| emotion type | non-harmful | harmful | emotion type | non-harmful | harmful | emotion type | non-harmful | harmful |
| gloom/sadness | 7 | 7 | joy | **32** | 12 | shame | 0 | **3** |
| relief | **18** | 9 | fondness | 21 | **49** | excitement | **5** | 3 |
| | | | dislike | 49 | **56** | fear | **11** | 3 |
| | | | | | | anger | 3 | **8** |
| | | | | | | surprize | **6** | 3 |
| SUM | **25** | 16 | SUM | 102 | **117** | SUM | **25** | 20 |

On the other hand, fondness scored unexpectedly higher in cyber-bullying dataset. Detailed analysis revealed that people would often express strong sarcasm with the use of positive expressions. Some examples of such entries have been represented in Table 1.

Finally, we compared tendencies in the annotated emotion types with regard to the two-dimensional affect space. The results are represented in Table 4. In the valence dimension, negative emotions were annotated most often on harmful data, and positive emotions, on non-harmful data, which is a reasonable and predictable result. The differences were not that obvious, however, after exclusion of fondness, which, as mentioned above, was often used mostly in sarcasm, the differences became clearer. The smallest difference was observed in emotion types which can be classified as both positive or negative, as their valence usually depends on the particular context.

As for the activation dimension, non-harmful data was annotated as more vivid in the groups of deeply activated and deeply deactivated emotions. On the other hand, harmful data was annotated more often on emotions with moderately activated emotions, which provides another proof for the thesis set on the beginning of this section, namely, that slandering is often expressed in non-emotional statements. Moreover, we were able to set the linguistic feature most distinguishable for cyber-bullying activities, namely, vulgarities. This is an important clue in the development of machine learning algorithm for cyber-bullying detection.

## 5 MACHINE LEARNING METHOD FOR CYBER-BULLYING DETECTION

In this section we describe a machine learning method developed to handle cyber-bullying activities. The method consists of several stages, including creation of a lexicon of vulgar, slanderous and abusive words, slanderous information detection module, ranking of the information according to the level of their harmfulness, and visualization of the harmful information. The system flow chart is represented on Figure 4. The creation of the system was separated into two general phases: training phase and processing (test) phase. Below we present the details of each phase.

1. Training phase:
   (a) Crawling the school Web sites,
   (b) Detecting manually cyber-bullying entries,
   (c) Extraction of vulgar words and adding them to lexicon,
   (d) Estimating word similarity with Levenshtein distance,
   (e) Part of speech analysis,
   (f) Training with SVM.

2. Processing (test) phase:
   (a) Crawling the school Web sites,
   (b) Detecting the cyber-bullying entries with SVM model,
   (c) Part of speech analysis of the detected harmful entry,
   (d) Estimating word similarity with Levenshtein distance,
   (e) Marking and visualisation of the key sentence.

### 5.1 Defining a Cyber-bullying Entry

Basing on the definition of cyber-bullying provided by MEXT [6] as well as the results of affect analysis from section 4, we created our own working definition of a cyber-bullying entry. With this definition we aimed to embrace the features needed to deal with in research such as ours.

Firstly, the data needs to be appropriately divided and weighted. In our definition we divided entries into three types. Normal (N) entry is an entry not containing any harmful information and thus not needing any intervention. Doubtful (D) entry contains information that might be harmful, but it needs further analysis to make the decision whether to delete it or not. Finally, Harmful (H) entry is a type of entry towards which there is no doubt that it contains harmful information and the decision to delete it can be made without further analysis. We used the above tripartite annotation to classify different types of entry contents.

The MEXT definition assumes that cyber-bullying happens when a person is personally offended on the Web. This includes disclosing the person's name, personal information and other areas of privacy. Therefore, as the first feature distinguishable for cyber-bullying entry we define NAMES. This includes such information as:

- **Names and surnames of people**
  (e.g. "Michal Ptaszynski")
  H - When a person's name can be clearly distinguished

- **Initials and nicknames**
  (e.g. "M.P.", "Mich** Ptasz**ski", "Mr. P.")
  H - When a person's identity can be clearly distinguished
  D - When a person's identity cannot be clearly distinguished

- **Names of institutions and affiliations**
  (e.g. "That JSPS Fellow from Hokkai-Gakuen University")
  H - When a person's identity can be clearly distinguished
  D - When a person's identity cannot be clearly distinguished

As the second type of a feature distinguishable for cyber-bullying we define any other type of PERSONAL INFORMATION. This includes:

- **Address, phone numbres, etc.**
  (e.g. "Minami 26, Nishi 11, Chuo-ku, Sapporo, 064-0926, Japan", or "+81-11-551-2951")
  H - When the information refers to a private person
  D - When the information is public or refers to a public entity

- **Questions about private persons**
  (e.g. "Who is that tall foreigner walking around lately on High-Tech corridor?")
  H - Always considered as undesirable and harmful, including situations, in which the object is described in a positive way

- **Entries revealing personal information**
  (e.g. "I heard that guy is responsible for the new project.")
  H - When a person's identity can be clearly distinguished
  D - When a person's identity cannot be clearly distinguished

Literature on cyber-bullying indicates vulgarities as the first distinctive feature of cyber-bullying activity [8, 9]. We were also able to prove this statement in the section 4. The biggest difference in extraction of linguistic features from both cyber-bullying and normal data was noticed for vulgarities. Therefore we selected VULGARITIES as the keywords distinguishable for cyber-bullying. Vulgarities are obscene or vulgar words which connote offences against particular persons or society. Examples of such words in English are *shit*, *fuck*, *bitch*. Examples in Japanese include words like *uzai* (freaking annoying), or *kimoi* (freaking ugly). In our research we divided the entries containing vulgarities into two types, namely:

- **Entries containing any form of vulgar or offensive language**
  D - Even when the object cannot be identified

- **Quarrels between two or more users**
  D - Even whet it happens between two anonymous users, a quarrel can lead to revealing personal information or other forms of cyber-bullying

All entries not containing any of the above information is classified as normal (N).

### 5.1.1 Definition Testing

To test, whether the definition is coherent, we performed an experiment. From the overall 2,999 entries (1,495 harmful and 1,504 normal) we extracted randomly a sample of 500 entries. Then we asked six human annotators (5 males and 1 female) to annotate the sample according to the above definition. Finally, we calculated Cohen's *kappa* agreement coefficient for all participants. The agreement coefficient was 0.67, which is regarded as strong, which confirms that the definition is coherent. However, in some cases the participants did not agree completely, mostly in cases of vulgarity. Some words could be considered as vulgar or not depending on one's subculture and usual vocabulary (e.g., "gangster", or "pimp" can be perceived as normal in a hip-hop subculture). This indicates that the definition should be complemented in the future with more strict definitions of vulgarities.

### 5.2 Construction of Vulgarity Lexicon

Vulgar words are often not detected by part of speech (POS) taggers, or marked as "unknown word". We decided to create list of vulgarities and add it to a basic lexicon of POS tagger. This was performed according to the procedure described below. At first we performed a study on the vulgar keywords constituting the harmful information. We obtained a set of informal school Web pages, read them and performed a manual categorization into "harmful" and "non-harmful" entries based on the MEXT classification of cyber-bullying. From these Web pages we obtained 1,495 harmful entries including 255 unique vulgar keywords distinguishable for cyber-bullying activity. The extracted keywords were finally added to the list of vulgarities. Finally, we added grammatical information to the extracted keywords and added them to the POS tagger[8]. An example of addition of grammatical information is presented in Table 5.

Table 5. Example of a newly registered word.

| |
|---|
| *kimoi* (freaking ugly) |
| POS: Adjective; |
| Headword: *kimoi* (hit-rate: 294); |
| Reading: kimoi; |
| Pronunciation: kimoi; |
| Conjugated form: uninflected; |

### 5.3 Estimation of Word Similarity with Levenshtein Distance

Users would often change spelling of words and write them in a unnormalized way. It is a part of jargonization of the language used online. Examples of this phenomenon in English, would be writing phrases like, "C.U." in the meaning of "See you [later]!" (usually on the end of e-mails or online messages), or "brah" in the meaning of "bro[ther], friend"[9], etc. Some examples of such colloquial transfor- mations in the Japanese online language are shown in Table 6.

---
[8] In this research we use standard POS tagger for Japanese, MeCab [17].
[9] http://www.internetslang.com/

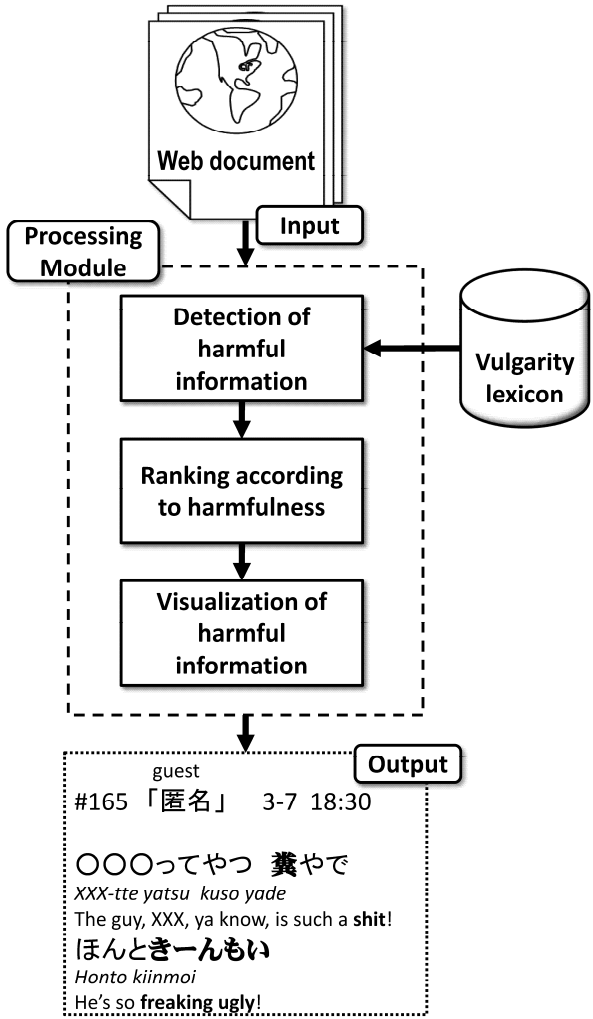

**Figure 4.** Flow chart of the cyber-bullying activity detecting system.

**Table 6.** Three examples of colloquial transformations of Japanese words in online jargon.

| original word | colloquial transformation |
|---|---|
| *kimoi* (freaking ugly, gross) | *kimosu, kishoi, kisho, ...* |
| *uzai* (freaking annoying) | *uzee, UZAI, uzakkoi, ...* |
| *busaiku* (ugly bitch) | *buchaiku, bussaiku, ...* |

With this variation, words having the same meaning would be classified as separate samples, which would cause hit-rate dispersion. Therefore to unify the same words written with slightly different spelling we calculated the similarity of the extracted words. The similarity was calculated using Levenshtein Distance [11] in a way similar to [12, 13]. The Levenshtein Distance between two strings is calculated as the minimum number of operations required to transform one string into another, where the available operations are only deletion, insertion or substitution of a single character.

However, as Japanese is transcribed using three character types, Chinese characters (*kanji*) encapsulating from one to several syllables and two additional syllabary (*katakana* and *hiragana*), calculating the distance would become imprecise. To solve this problem, every word was automatically transformed in their alphabetical transcription. An example of distance calculation is presented on the back-transformation of the word *kimosu* to its original spelling *kimoi* in Table 7. The distance between both words is equal 2.

**Table 7.** Three examples of colloquial transformations of Japanese words in online jargon.

| transformed word | performed operation |
|---|---|
| *kimosu* | |
| → *kimoiu* | substitution of 's' to 'i'; distance = 1; |
| → *kimoi* | deletion of final 'u'; distance = 2; |

### 5.4 SVM Based Classification of Cyber-bullying

To classify the entries into either harmful (cyber-bullying) or non-harmful, we used Support Vector Machines. Support Vector Machines (SVMs) are a method of supervised machine learning developed by Vapnik [14] and used for classification of data. They are defined as follows. With a set of training samples, divided into two categories A and B, SVM training algorithm generates a model for prediction of whether test samples belong to either category A or B. In the traditional description of SVM model, samples are represented as points in space (vectors). SVM constructs a hyperplane, in a space of a higher dimension than the base one, with the largest distance to the nearest training data points (support vectors). The larger the margin the lower the generalization error of the classifier. Since SVM has been successfully used for text classification [15] we decide to use them in this research as well. In our research the category A contains cyber-bullying cases and the category B contains all other cases, which do not consist of socially harmful information. As the software for building SVM models we used SVM light (ver6.02)[10].

### 5.5 Extraction of Key Sentences

In the process of automation of Online Patrol, apart from the classification of cyber-bullying entries, there is a need to appropriately determine how harmful is a certain entry. A ranking according to the harmfulness of entries is important to detect the most dangerous cases. In our approach an entry is considered as the more harmful, the more vulgar keywords appear in the entry.

The harmfulness of an entry is calculated using T-score. T-score is a measure that answers the question of how confident one can be that the association measured between two words is an actual collocation and not a matter of chance. The higher occurrence frequency a word has in a corpus, the higher is the value of T-score. A T-score of a word associating with words A and B is calculated according to the equation below:

$$T_{score} = \frac{a}{b} \quad (1)$$

where,

$$a = [word\,co-occurrence\,frequency] - \frac{([occurence\,of\,word\,A] * [occurence\,of\,word\,B])}{[all\,words\,in\,the\,corpus]}$$

and,

---
[10] http://svmlight.joachims.org

$$b = \sqrt{[word\,co-occurrence\,frequency]}$$

We calculate the harmfulness of the whole entry as a sum of T-scores calculated for all vulgar words. This way the more frequently occurring words there are in the entry, the higher rank the entry achieves in the ranking of harmfulness.

## 6 EVALUATION OF THE METHOD

To verify the performance of the method we evaluated three procedures:

1. Classification of Cyber-bullying entries with SVM,
2. Word similarity calculation with Levenshtein distance,
3. Extraction of key sentences.

### 6.1 Evaluation of SVM Model

To apply SVM to detect harmful information from unofficial school BBS sites, we needed to prepare the data for training the SVM model. At first we performed morphological analysis of the BBS entries to be used as the training data. For every part of speech from the analyzed and parsed data we used as features the POS labels with the original strings of characters. As the features in the part of speech identification we used parts of speech like nouns (person's name, or other than name), verbs and adjectives. In the identification of the whole entries we used feature sets consisting of the features of each part of speech and the strings of characters containing the whole entry. Based on the amount of identified features, SVM model calculates the probability of affiliation of a character string to a certain class.

As the features for training we used several combinations of main and additional features. Main features included: (1) words with POS, (2) words only, (3) POS only. Additional features included: (A) Occurrence frequency, calculated as in equation 2, (B) Relative frequency, calculated as in equation 3, (C) Inverse document frequency (IDF), calculated as in equation 4, and (D) Term frequency–inverse document frequency (TF-IDF), calculated as in equation 5.

$$(A)\ Occurrence\,frequency = Frequency\,of\,a\,POS\,within\,one\,document \quad (2)$$

$$(B)\ Relative\,frequency = \frac{(A)}{Frequency\,of\,the\,POS\,within\,all\,documents} \quad (3)$$

$$(C)\ IDF = \log(\frac{number\,of\,all\,entries}{number\,of\,entries\,containing\,the\,POS} + 1) \quad (4)$$

$$(D)\ TF-IDF = (A) * (C) \quad (5)$$

As the training data we used all 2,999 entries gathered during an actual Online Patrol, from which human annotators (Online Patrol members) classified 1,495 entries as harmful and 1,504 as non-harmful. We did not, however, apply the string similarity calculation to the SVM model, to evaluate both techniques separately.

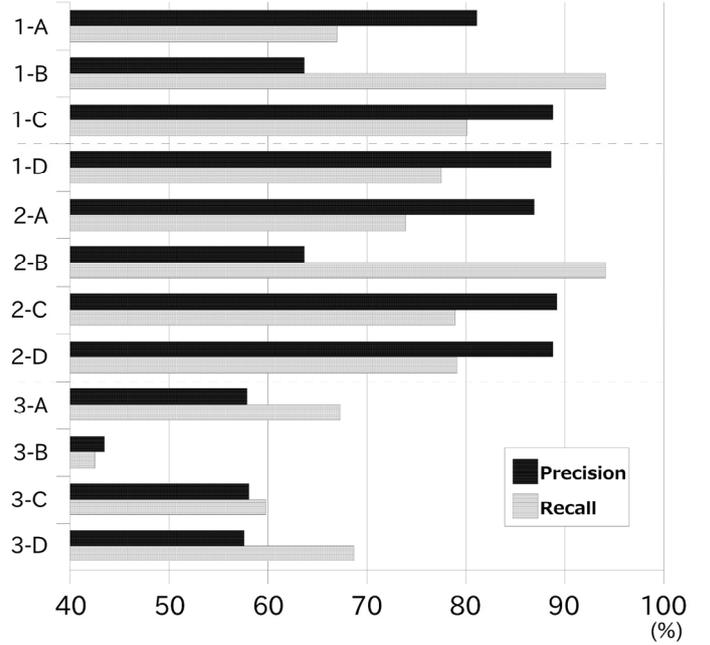

**Figure 5.** Results of the experiments with SVM for different features.

The above conditions are applied to test the model trained with SVM_light. In the evaluation we calculated the system's result as balanced F-score, using 10-fold cross validation for Precision and Recall. In 10-fold cross validation data is first broken into 10 sets of size n/10. Then, 9 datasets are used to train on and 1 as a test. This procedure is repeated 10 times and the overall score is the mean accuracy from all 10 tests. The balanced F-score, Precision (P) and Recall (R) are calculated as follows,

$$F_{score} = 2 * \frac{P * R}{P + R} \quad (6)$$

where,

$$Precision = \frac{s}{n} \quad Recall = \frac{n}{c}$$

and,
$s$ = cases correctly classified by the system as harmful
$n$ = all cases classified by the system as harmful
$c$ = all harmful cases

The results are represented in Figure 5. The results for group 3 (only POS) were the lowest. Groups 1 (words+POS) and 2 (words only) were comparable. The highest Recall appeared in the experiments with Relative Frequency. However, the Precision in this case was not ideal (close to 65%). Experiments with IDF (C) and TF-IDF (D) gave the highest results with Recall close to 80%, Precision close to 90%. The highest score from all was for the combination words+POS with TF-IDF (Precision = 89%, Recall = 80%, F-score = 84.3%).

### 6.2 Evaluation of Similarity Calculation

When preparing the conditions for evaluation of word similarity calculation method with Levenshtein distance, we noticed that the larger was the threshold, the higher was the probability of matching a word with completely different meaning. Therefore we performed an optimization of the similarity calculation algorithm. In the optimization we applied two heuristic rules shown in Table 8.

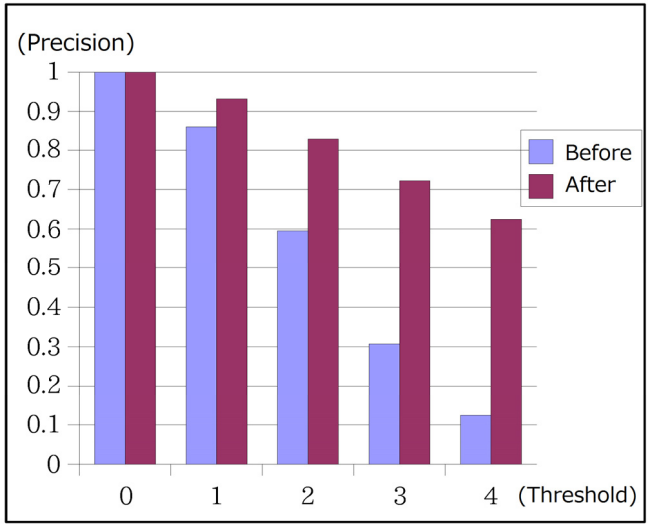

**Figure 6.** Precision of similarity calculation before and after applying the heuristic rules.

**Table 8.** Two heuristic rules applied in optimization of similarity calculation algorithm.

| Rule | Example |
|---|---|
| 1. deletion syllable prolongations | *kimoooi* → *kimoi* |
| 2. unification of word first letter | In case of *uzai* we will consider only the words beginning with *u* |

On BBS sites, where informal language is widely used, such as unofficial school Web sites, prolonging of syllables is used mostly to express changes in user attitude, or mood, usually highlighting the unofficial character of the entry. However, such operation has no influence on the semantics of a word. Therefore before the phase of similarity calculation we added a rule deleting all syllable prolongations.

The second rule was unification of the first letters of matched words. This was done because in the Japanese language only the final letters change during conjugation. Therefore we could assume that it is irrelevant to calculate similarity for words differing with first letters.

The results for Precision of similarity calculation before and after applying the heuristic rules are presented on Figure 6. The Precision was greatly improved after applying the rules. With the threshold set on 2, the Precision before applying the rules was 58.9% and was improved to 85.0%.

### 6.3 Evaluation of Key Sentence Extraction

To set the ranking of cyber-bullying entries according to how harmful they are, we calculated the T-score of vulgar vocabulary they contain. The results are represented in Table 9. As one can see in Table 9, the first three scores are much higher then the other ones. This is caused by numerous repetitions of the same vulgar words in one entry, which artificially increases the occurrence rate. Moreover, in the cases, appearing in the table on places 4-6, one of the words was used repeatedly, which also caused artificial increase of occurrence and eventually the result of T-score, although smaller than for the first three cases. In such cases, although the occurrence of a certain pair is not high, T-score is artificially increased and therefore biased.

To solve this problem we considered the same vulgar words appearing in one entry as one word. However, some vulgar words may appear as collocations. In such cases the identical numerous collocations are considered as one. The change in the results is represented in Table 10. The results indicate the bias have disappeared. However, the T-score became too small causing many word pairs appear on the same place in the ranking making the ranks difficult to set. This is caused by the fact that the words with the same meaning appear on the Web sites in different transcription and therefore there is a large number of word sets with a small occurrence.

We solved this problem by calculating word similarity before calculating T-score. The results are represented in Table 11. Due to word similarity calculation a rise in T-score was observed and the number of diversified word pairs increased, which made rank setting much easier. There was 202 vulgar word pairs, from which 40 pairs occurred more than twice.

**Table 9.** The results of T-score calculation for sets of vulgar words. As mentioned in section 5.3 words in Japanese can be usually transcribed in three different systems: *hiragana*, *katakana* and *kanji*. The differences in transcription are represented in the table as markers after the words, with [h] for *hiragana*, [k] *katakana* and [K] for *kanji*.

| word A | word B | co-occurrence | T-score |
|---|---|---|---|
| *baka* [h] (stupid) | *baka* [h] | 861 | 29.34 |
| *shine* [K] (fuck you) | *shine* [K] | 552 | 23.49 |
| *shine* [h] | *shine* [h] | 58 | 7.62 |
| *tarashi* [k] (pimp) | *shine* [K] | 17 | 4.12 |
| *busu* [k] (ugly bitch) | *shine* [K] | 16 | 4.00 |
| *kimosu* [h] | *shine* [h] | 16 | 4.00 |
| *shine* [h] | *busu* [h] | 6 | 2.45 |

**Table 10.** Change in the results of T-score calculation for sets of vulgar words, when two or more words were considered as one.

| word A | word B | co-occurrence | T-score |
|---|---|---|---|
| *shine* [K] | *shine* [K] | 7 | 2.65 |
| *shine* [h] | *shine* [h] | 4 | 2.00 |
| *pashiri* [k] (looser) | *shine* [K] | 3 | 1.73 |
| *debu* [k] | *kiero* [K] | 3 | 1.73 |
| *kiero* [K] (get lost) | *kiero* [K] | 2 | 1.41 |
| *shine* [K] | *kiero* [K] | 2 | 1.41 |
| *uzai* [h] | *kimoi* [k] | 2 | 1.41 |

**Table 11.** Change in the results of T-score calculation for sets of vulgar words, with calculated word similarity.

| word A | word B | co-occurrence | T-score |
|---|---|---|---|
| *shine* [K] | *shine* [K] | 11 | 3.32 |
| *kimoi* [h] | *shine* [K] | 11 | 3.32 |
| *kimoi* [h] | *busaiku* [h] | 8 | 2.83 |
| *uzai* [h] | *kimoi* [h] | 7 | 2.65 |
| *panko* [k] (slut) | *panko* [k] | 6 | 2.45 |
| *kimoi* [h] | *kimoi* [h] | 6 | 2.45 |
| *busaiku* [h] | *busaiku* [h] | 6 | 2.45 |

### 6.4 Discussion

This time in morphological analysis of vulgar words we used only a small set of manually discovered words, which we added to the lexi-con. It is difficult to include all existing vulgar words and more such vocabulary will appear in the future as well. Therefore as a further work we need to develop a method for automatic extraction of vulgar vocabulary from the Internet.

The results of SVM model used to distinguish between harmful and non-harmful information were 89% of Precision and 80% of Recall. However, on the unofficial school Web pages used as the data in this research there were numerous entries consisting of only one sentence, or even one word. Therefore the feature set for training was not sufficient and the overall result was not ideal (F-score = 84.3%).

As for word similarity calculation, many vulgar words are short and a change of even one letter might cause a change of meaning. Therefore, two different words would be matched as similar by Levenshtein distance, when the threshold is too wide. This problem might be solved by automatically setting the threshold according to the word length.

As for the extraction of key sentences, although we were able to calculate a non biased T-score for vulgar expressions and set the ranking, over 80% of vulgar words appeared only once. This caused over half of the cases to be attached with similar ranks. This problem could be solved by increasing the number of training data or applying a different method of rank setting.

## 7 CONCLUSIONS AND FUTURE WORK

In this paper we presented a research on cyber-bullying, a new social problem that emerged recently, together with the development of social networking portals, etc. Cyber-bullying consist of sending messages containing slanderous expressions, harmful for other people, or verbally bullying other people in front of the rest of online community. In Japanese society, on which we focused in particular, this problem is particularly vivid on unofficial school Web sites. To handle the problem, teachers and PTA members perform voluntarily Online Patrol to spot and delete the online entries harmful for other people. Unfortunately, there is already an enormous number of cyber-bullying cases and the number keeps growing, which makes the Online Patrol an uphill task when performed manually.

We started this research to create an artificial Online Patrol agent. Looking for clues to select linguistic features to be used in machine learning algorithm, we performed comparative affect analysis of the cyber-bullying data and normal entries. As a result, we noticed that, that the harmful data were less emotively emphasized than non-harmful. The thesis is reasonable, since the harmful entries are written with premeditation and aim not in expressing ones own emotions, but in evoking in other online community members negative emotions against victim of the cyber-bullying. The results of comparing different dimensions of emotional emphasis suggested the thesis was true, although the reliability of the proof was not satisfactory and further analysis on more robust data is necessary. Another discovery, although an expected one, was that positive emotions appeared more often in non-harmful data and negative emotions appeared more often in harmful data. However, detailed analysis revealed that, especially for fondness, the expressions of positive emotions are often used in strong sarcastic meaning. Therefore there is a need to analyze the data taking into consideration also other dimensions than the valence and the activation of an emotion. As one of such means we plan to apply Ptaszynski et al's [25, 26] system for contextual affect analysis verifying whether an emotion expressed in an utterance is appropriate for its context. We assume this will help in developing a sufficient model of formalization of cyber-bullying activities. For the first step of the research, we were able to select the most distinctive linguistic feature for cyber-bullying, namely, vulgarities. We used this feature in the creation of machine learning classifier for cyber-bullying detection.

We created the machine learning-based system for cyber-bullying detection and evaluated it. At first, we manually gathered a lexicon of vulgar words distinctive for cyber-bullying entries. To recognize the vulgar words, but written in an informal or jargonized way, we calculated word similarity with Levenshtein distance. As the result, with the threshold equal 2 the system was able to correctly determine the similar words with a 85% of Precision. The Support Vector Machine model trained for discrimination between harmful and non-harmful data achieved the results of 89% of Precision and 80% of Recall (with balanced F-score = 84.3%). Finally the rank setting of the online entries according to their harmfulness was set using T-score. We were able to eliminate undesirable bias in the rank setting, unfortunately, the procedure is not yet ideal.

Since new vulgar words appear frequently, we need to find a way to automatically extract new vulgarities from the Internet to keep the lexicon up to date. There is also a need for more experiments with the system including its different variations and improvements (ex. Levenshtein distance threshold optimization).

The problems concerning online security have been escalating ever since the birth of the Internet. Some of them are widely known to the society, such as spam e-mails or hacking, however more and more such problems, including cyber-bullying, appear by the day and social consciousness about them is not yet sufficient. There have been developed solutions for some of these problems (e.g., automatic spam e-mail detection, firewall protection, etc.), while others, like cyber-bullying, only keep escalating. Recognising such problems and developing the remedy is an urgent matter and is desirable to come into focus of Artificial Intelligence.

## ACKNOWLEDGEMENTS

This research was supported by (JSPS) KAKENHI Grant-in-Aid for JSPS Fellows (Project number: 22-00358) and a Research Project on Development of a Internet Word Book System for Dynamic Following of Information Transition (Project number: 20500833). The authors thank Mr. Motoki Matsumura from Human Rights Research Institute Against All Forms for Discrimination and Racism-MIE for providing data from unofficial school Web sites.